\documentclass{article}

\usepackage[final]{neurips_2025}

\usepackage[utf8]{inputenc} % allow utf-8 input
\usepackage[T1]{fontenc}    % use 8-bit T1 fonts
\usepackage{hyperref}       % hyperlinks
\usepackage{url}            % simple URL typesetting
\usepackage{booktabs}       % professional-quality tables
\usepackage{amsfonts}       % blackboard math symbols
\usepackage{nicefrac}       % compact symbols for 1/2, etc.
\usepackage{microtype}      % microtypography
\usepackage{xcolor}         % colors

\usepackage{graphicx}
\usepackage{subcaption}
\usepackage{amsmath,amsthm,amssymb}
\usepackage{booktabs}
\usepackage{caption} 
\title{Random Projection Flows for Efficient Manifold Density Estimation}

\author{%
  Ahmad Ayaz Amin \\
  Department of Computer Science\\
  Toronto Metropolitan University\\
  Toronto, Canada \\
  \texttt{ahmadayaz.amin@torontomu.ca}
  \And
  Baha Uddin Kazi \\
  Faculty of Applied Science and Technology\\
  Humber Polytechnic\\
  Toronto, Canada \\
  \texttt{bahauddin.kazi@humber.ca}
}

\begin{document}

\maketitle

\begin{abstract}
Accurate density estimation is crucial for understanding complex high-dimensional data, but it becomes challenging when the data lies on or near low-dimensional manifolds. Random projections provide a natural way to reduce dimensionality while approximately preserving geometric structure, enabling effective density estimation in these settings. We introduce \emph{Random Projection Flows} (RPFs), a principled framework for injective normalizing flows that leverages tools from random matrix theory and the geometry of random projections. RPFs employ random semi-orthogonal matrices, drawn from Haar-distributed orthogonal ensembles via QR decomposition of Gaussian matrices, to project data into lower-dimensional latent spaces for the base distribution. Unlike principal component analysis flows or learned injective maps, RPFs are plug-and-play, efficient, and yield closed-form expressions for the Riemannian volume correction term. We demonstrate that RPFs are both theoretically grounded and practically effective, providing a strong baseline for generative modeling and a bridge between random projection theory and normalizing flows.
\end{abstract}

\section{Introduction}
Normalizing flows \citep{pmlr-v37-rezende15, DBLP:journals/corr/DinhKB14} are a class of generative models that provide exact likelihood evaluation and flexible density estimation by transforming a simple base distribution (typically Gaussian) into a complex data distribution via a sequence of invertible transformations. Formally, given a data sample $\mathbf{x} \in \mathbb{R}^D$ and an invertible transformation $f_\theta$ with parameters $\theta$, the change-of-variables formula gives the log-likelihood:
\begin{equation}
    \log p_\theta(\mathbf{x}) = \log p_Z(f_\theta(\mathbf{x})) + \log \left| \det \frac{\partial f_\theta(\mathbf{x})}{\partial \mathbf{x}} \right|
\end{equation}
where $p_Z$ is the density of the base distribution and the second term is the log determinant of the Jacobian.  

Although fully invertible flows can be made arbitrarily flexible \citep{grathwohl2019ffjord}, high-dimensional data often motivate the use of \emph{injective flows}, which map data into a lower-dimensional latent space $d < D$. For injective flows \citep{gemici2016normalizingflowsriemannianmanifolds, caterini2021rectangular}, the log-likelihood must account for the dimensionality reduction:
\begin{equation}
    \log p_\theta(\mathbf{x}) = \log p_Z(f_\theta(\mathbf{x})) - \frac{1}{2} \log \left| \det \left(J_\theta(f_\theta(\mathbf{x}))^\top J_\theta(f_\theta(\mathbf{x}))\right) \right|
\end{equation}
where $J_\theta(\mathbf{x})$ is the $D \times d$ Jacobian of the inverse of the injective transformation. A central difficulty is computing the Riemannian volume element, as it involves determinants and matrix inverses. 

Although dimensionality reduction for density estimation via random projections is a classic idea, the design of random projections explicitly tailored for density estimation back in the ambient space, with guarantees of correctness and scalability, has not been deeply explored. Our approach makes this explicit: rather than learning a subspace, we fix a Haar-random projection \citep{mezzadri2006generate}. Thanks to the semi-orthogonal nature of the projection matrix, the corresponding Jacobian volume term has a closed form. An optional constant scaling factor motivated by Johnson--Lindenstrauss (JL) theory \citep{johnson1984extensions} can be applied to make projected norms unbiased in expectation, but this is not required for correctness.

Importantly, the JL-inspired scaling is best viewed as a convenient add-on: one could omit it (yielding a strictly isometric map onto the random subspace), or even replace it with a learned scale as part of the model. We show empirically that including the JL-motivated constant often improves likelihood calibration and density estimates compared to PCA flows, but the random projection flow itself remains valid regardless.

Our contributions are:
\begin{enumerate}
  \item We propose Random Projection Flows (RPFs), a class of injective flows using Haar-distributed semi-orthogonal projections with a simple, optional JL-motivated volume correction.
  \item We analyze this correction: although not necessary, it provides a constant closed-form Jacobian term and can improve calibration; it could also be dropped or learned.
  \item We position RPFs conceptually between two-stage density estimators (fixed encoder + separate density model) and fully end-to-end latent-variable models, enjoying benefits of both while avoiding manifold overfitting common in VAEs.
  \item We demonstrate that RPFs are composable: they can be used standalone (e.g. with a GMM) or as building blocks inside arbitrary normalizing flow or SurVAE \citep{NEURIPS2020_9578a63f} architectures while preserving exact likelihoods.
  \item Empirically, RPFs outperform PCA-based injective flows on UC Irvine (UCI) benchmarks and preserve manifold geometry in synthetic experiments.
\end{enumerate}

\section{Random Projections and Random Matrix Theory}
\subsection{Haar Orthogonal Matrices from Gaussian QR}
A standard method to generate a Haar-distributed orthogonal matrix $Q \in \mathbb{R}^{D \times D}$ is to sample a Gaussian matrix $G$ with i.i.d. $\mathcal{N}(0,1)$ entries and apply QR decomposition $G = QR$. The orthogonal factor $Q$ is Haar distributed \citep{mezzadri2006generate}. Taking the first $d$ rows of $Q$ yields a semi-orthogonal matrix $V \in \mathbb{R}^{d \times D}$ with $VV^\top = I_d$. This induces a uniform distribution over $d$-dimensional subspaces (the Grassmannian).

\subsection{The Johnson–Lindenstrauss (JL) Lemma}
The JL lemma \citep{johnson1984extensions} provides a theoretical foundation for dimension reduction via random linear projections. It states that a set of $N$ points $\{x_i\} \subset \mathbb{R}^D$ can be embedded into $\mathbb{R}^d$ with $d = O(\epsilon^{-2} \log N)$ such that all pairwise distances are approximately preserved:
\begin{equation}
(1-\epsilon) \|x_i - x_j\|_2 \le \|R x_i - R x_j\|_2 \le (1+\epsilon) \|x_i - x_j\|_2, \quad \forall i,j.
\end{equation}
Gaussian and Haar-distributed semi-orthogonal matrices satisfy this property with high probability. These embeddings are approximate isometries, with a scaling factor that concentrates tightly; this underpins our choice of volume correction in RPFs.

\subsection{Singular Value Concentration}
The singular values of random rectangular Gaussian matrices follow the Marchenko--Pastur law \citep{marchenko1967distribution}, implying concentration of $\|Rx\|_2^2$ around its expectation. This explains both the JL distance-preservation guarantees and why the Jacobian volume term in RPFs reduces to a simple constant.

\section{Random Projection Flows}

We introduce \emph{Random Projection Flows (RPF)}, an efficient class of injective flows that leverage random linear projections to compress high-dimensional data into a lower-dimensional latent space, followed by a tractable latent density model such as a Gaussian mixture model (GMM).

\subsection{Random Linear Projection and Volume Element}
Let $\mathbf{x} \in \mathbb{R}^D$ and $V \in \mathbb{R}^{d \times D}$ be Haar semi-orthogonal with $VV^\top = I_d$. We define a scaled projection:
\begin{equation}
W = \sqrt{\frac{D}{d}}\, V, \qquad \mathbf{z} = W \mathbf{x}.
\end{equation}
This JL-inspired scaling keeps norms approximately unbiased. The Jacobian is $J=W$ and the Riemannian metric is
\begin{equation}
J^\top J = W^\top W = \frac{D}{d} V^\top V.
\end{equation}
$V^\top V$ projects onto a $d$-dimensional subspace with $d$ eigenvalues $1$ and $D-d$ zeros. Hence
\begin{equation}
\det(J^\top J) = \left(\frac{D}{d}\right)^d, \quad 
\sqrt{\det(J^\top J)} = \left(\frac{D}{d}\right)^{d/2},
\end{equation}
yielding a constant log-volume correction:
\begin{equation}
\log \sqrt{\det(J^\top J)} = \frac{d}{2} \log \frac{D}{d}.
\end{equation}
This constant appears with opposite signs when encoding versus decoding, reflecting the symmetry of the change-of-variables formula. One could drop it (making the map strictly isometric) or even learn the scale along with the shift, resulting in an injective analogue of the affine flow in RealNVP \citep{dinh2017density} and Glow \citep{NEURIPS2018_d139db6a}. We do not pursue the latter, leaving that for future work, but we do show that the JL-derived constant can yield better results than the fully isometric model in certain scenarios (and vice versa).

\subsection{Latent Density Modeling and Positioning}
In the latent space, we fit a tractable density such as a GMM:
\[
p_Z(\mathbf{z}) = \sum_{k=1}^K \pi_k \mathcal{N}(\mathbf{z}; \mu_k, \Sigma_k).
\]
Then, the data log-likelihood is
\begin{equation}
\log p_X(\mathbf{x}) = \log p_Z(W \mathbf{x}) + \frac{d}{2}\log\frac{D}{d}.
\end{equation}
RPFs thus sit conceptually between classic two-stage methods (PCA+GMM) and end-to-end latent-variable models. Like the former, the projection is fixed; like the latter, the whole map is differentiable with exact likelihood. RPFs avoid manifold overfitting \citep{loaiza-ganem2022diagnosing} and post-hoc density correction \citep{dai2018diagnosing} often needed in variational autoencoders \citep{Kingma2013AutoEncodingVB} and other maximum likelihood models with manifold latent dimensionality.

Furthermore, because the projection has a closed-form volume element, RPF layers and their trainable counterparts can be dropped into arbitrary normalizing flow or SurVAE \citep{NEURIPS2020_9578a63f} architectures while preserving exact likelihoods.

\subsection{Properties}
The key properties of RPFs are:
\begin{itemize}
\item \textbf{Injectivity:} $W$ is injective almost surely.
\item \textbf{Constant volume term:} additive constant, independent of $x$.
\item \textbf{JL guarantees:} high-probability distance preservation; scaling matches expected norms.
\item \textbf{Haar invariance:} rotationally unbiased.
\end{itemize}

Although having many benefits, RPFs do have a key disadvantage in being unable to scale to very complex datasets, which we show in the CIFAR-10 experiment.

\section{Related Works}
Normalizing flows \citep{pmlr-v37-rezende15, DBLP:journals/corr/DinhKB14} provide invertible maps with tractable likelihoods. Injective extensions \citep{gemici2016normalizingflowsriemannianmanifolds, caterini2021rectangular} handle dimensionality reduction but incur expensive Riemannian volume terms. PCA flows \citep{cramer2022principalcomponentdensityestimation} exploit semi-orthogonal projections learned from data. RPFs generalize these by using random Haar projections: unbiased across subspaces and with closed-form volume correction.

Random projection theory underpins RPFs. The JL lemma \citep{johnson1984extensions, dasgupta2003elementary, achlioptas2003database, vempala2005random} shows Gaussian and sub-Gaussian maps preserve distances. Haar matrices from Gaussian QR \citep{mezzadri2006how} have well-characterized spectra \citep{edelman2005random, diaconis1994eigenvalues}. Concentration results \citep{vershynin2018highdimensional} explain the near-isometry.

Structured orthogonal embeddings further motivate RPFs. Orthogonal Random Features \citep{NIPS2016_53adaf49} reduce kernel approximation variance using random orthogonal matrices. Felix et al. \citep{NIPS2017_bf822969} demonstrate the “unreasonable effectiveness” of structured embeddings (e.g. Hadamard), achieving JL-like guarantees at lower cost. These lines of work show random orthogonal maps are powerful primitives across ML; RPFs leverage them for generative modeling with exact likelihoods.

\section{Experiments}

\subsection{UCI Density Esimation}
We evaluate RPFs on the UCI density estimation benchmarks introduced in \citep{NIPS2017_6c1da886}, including \texttt{POWER}, \texttt{GAS}, \texttt{HEPMASS}, and \texttt{MINIBOONE}. These datasets are widely used to assess the quality of density estimators under moderate-dimensional structured data.

\subsubsection{Setup}
Following standard practice, we preprocess the datasets using the protocol of \citep{NIPS2017_6c1da886} and split them into training, validation, and test sets.  

We compare RPFs against flows constructed with PCA projections \citep{cramer2022principalcomponentdensityestimation}, which serve as a strong baseline for injective flow layers. Each method is combined with a Gaussian mixture model (GMM) base density. For fair comparison, we train all models under identical optimization settings, including the number of manifold dimensions and the number of Gaussian mixture components, and report test log likelihoods. The results are shown in Table \ref{tab:uci_density_extended}.

\subsubsection{Results}
Table~\ref{tab:uci_density_extended} reports test log-likelihoods on the UCI benchmark datasets for three models: JL-scaled Random Projection Flows (JL), isometric RPFs without the JL scaling factor (ISO), and PCA-based injective flows.

Both RPF variants clearly outperform PCA across all datasets, often by several nats, despite using data-independent random projections. This highlights their competitiveness as low-cost, data-agnostic alternatives. The JL scaling offers a small, dataset-dependent benefit, but is not essential as ISO performs similarly well. The key advantage is that even fixed random projections provide strong density estimates without the need for learned projections as in PCA.

\begin{table}[h]
\centering
\caption{Density estimation results (test log-likelihood) on UCI datasets. JL denotes RPFs with Johnson--Lindenstrauss-inspired scaling, ISO denotes isometric RPFs (no scaling), and PCA denotes PCA-based injective flows. Higher is better.}
\begin{tabular}{lcccc}
\toprule
Model & POWER & GAS & HEPMASS & MINIBOONE \\
\midrule
RPF (JL) & $-1.72 \pm 0.14$ & $-1.57 \pm 0.37$ & $-20.08 \pm 0.14$ & $-14.68 \pm 0.30$ \\
RPF (ISO) & $-1.99 \pm 0.21$ & $-1.40 \pm 0.04$ & $-19.97 \pm 0.20$ & $-14.63 \pm 0.47$ \\
PCA Flow \citep{cramer2022principalcomponentdensityestimation} & $-2.51 \pm 0.13$ & $-2.32 \pm 0.04$ & $-20.71 \pm 0.38$ & $-20.66 \pm 0.05$ \\
\bottomrule
\end{tabular}
\label{tab:uci_density_extended}
\end{table}

\subsection{Projection Comparison}
In addition to quantitative evaluation on synthetic 2D datasets, we investigated how RPFs and PCA behave on higher-dimensional data. Specifically, we examined three benchmark 3D datasets from scikit-learn \citep{scikit-learn}: the Swiss roll, the S-curve, and clustered blobs.

\subsubsection{Setup}
\begin{figure}[h]
    \centering
    \includegraphics[width=0.75\linewidth]{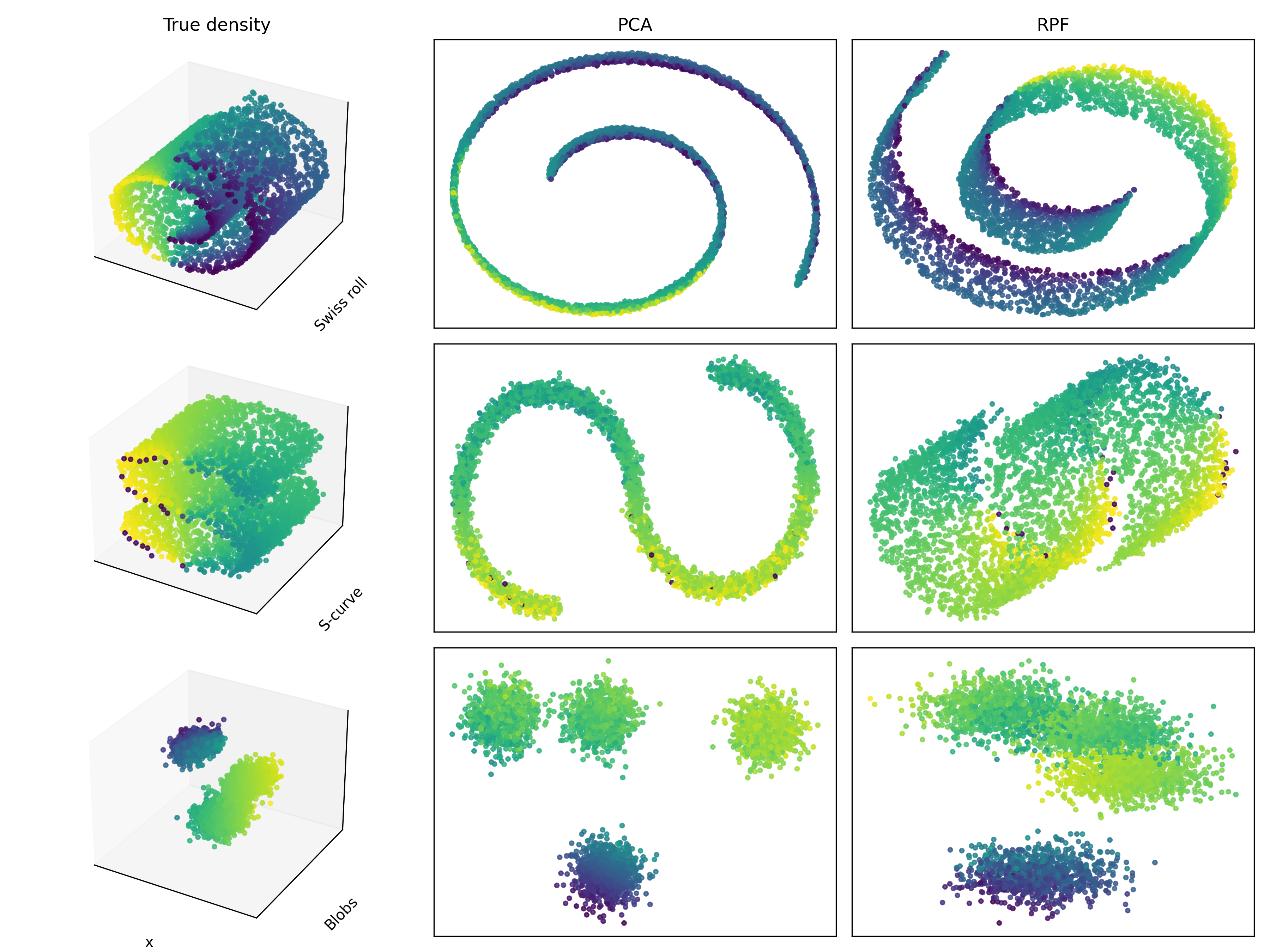}
    \caption{Projections of 3D benchmark datasets. Left: true 3D density with color gradients. Middle: PCA projection. Right: RPF projection.}
    \label{fig:sklearn_projections}
\end{figure}

For each dataset, we first display the true distribution in 3D, with a color gradient assigned according to one coordinate axis. This coloring allows us to track local structure after projection. We then visualize the corresponding 2D embeddings produced by PCA and RPF. The results are shown in Figure~\ref{fig:sklearn_projections}.

\subsubsection{Results}
\paragraph{Swiss roll} PCA flattens the manifold into a nearly linear band, losing the roll’s spiral geometry. In contrast, the RPF retains much of the nonlinear curvature, producing a projection that still resembles the original manifold.  

\paragraph{S-curve} PCA reduces the manifold to a simple arc, obscuring the double-banded shape. RPF better preserves the two-layered structure, suggesting a stronger correspondence with the underlying geometry.  

\paragraph{Blobs} Both methods maintain separability between clusters, but PCA aligns the blobs along its principal axes, while RPF introduces greater variability in density and orientation, highlighting differences in local structure.  

\paragraph{Implications for density estimation.} 
These projections illustrate why RPFs outperform PCA in UCI data. PCA collapses nonlinear manifold features by aligning with global variance, as seen in the Swiss roll and S-curve. Random orthogonal projections instead preserve local geometry more uniformly in expectation, leading to embeddings that better respect intrinsic structure — crucial for accurate likelihood estimation.

\subsection{MNIST and CIFAR-10 Density Estimation}

\subsubsection{Setup}
We perform unconditional density estimation on the MNIST and CIFAR-10 datasets, following the preprocessing methodology of \citep{NIPS2017_6c1da886}. We report log-likelihoods in logit space and compare to the results presented in \citep{NIPS2017_6c1da886}. 

We train a Random Projection Flow (RPF) with JL scaling, using a Gaussian Restricted Boltzmann Machine (GRBM \citep{liao2022gaussianbernoullirbmstears}) prior trained via persistent contrastive divergence (PCD \citep{tieleman2008training}). Log-likelihoods are estimated using annealed importance sampling (AIS \citep{neal1998annealedimportancesampling}) to approximate the partition function.

\subsubsection{Results}
As shown in Table~\ref{tab:mnist_density_extended}, the RPF model substantially outperforms MADE \citep{pmlr-v37-germain15} on MNIST. On CIFAR-10, however, RPF performs worse than MAF \citep{NIPS2017_6c1da886} and the Gaussian model. This may reflect the difficulty of modeling projected densities with a GRBM for complex, high-dimensional natural images. While MADE also underperforms the Gaussian model, it suggests that RPFs combined with GRBMs are effective for simpler datasets like MNIST but face limitations on CIFAR-10, likely due to the need for more expressive latent models or deeper architectures.

\begin{table}[h]
\centering
\caption{Unconditional density estimation results (test log-likelihood) on MNIST and CIFAR-10. RPF denotes RPFs with JL scaling. Higher is better; best-performing model is highlighted in \textbf{bold}.}
\begin{tabular}{lcc}
\toprule
Model & MNIST & CIFAR-10 \\
\midrule
RPF (ours) & $\mathbf{-149.7}$ & $-4885$ \\
\midrule
Gaussian &  $-1366.9$ & $2367$ \\
MADE MoG & $-1038.5$ & $-397$ \\
MAF (10) & $-1313.1$ & $\mathbf{3049}$ \\
\bottomrule
\end{tabular}
\label{tab:mnist_density_extended}
\end{table}

%\subsection{Discussion}
%RPF’s plug-and-play nature allows easy integration into existing flow architectures, unlike PCA or learned injective transformations. It offers a compact, efficient alternative for density estimation, particularly in small-data regimes. RPFs are a practical tool for prototyping before committing to full learned injective flows.

\section{Conclusion}

We introduced \emph{Random Projection Flows (RPFs)}, a simple and efficient class of injective normalizing flows that leverage Haar-distributed semi-orthogonal projections to compress high-dimensional data into a lower-dimensional latent space. By exploiting the constant volume element of these projections, RPFs provide tractable likelihood evaluation without per-sample determinant computations. 

Empirically, RPFs achieve competitive density estimation performance on UCI benchmarks and synthetic manifolds, often outperforming PCA-based injective flows despite being entirely data-independent. Our experiments highlight that the geometry preserved by random projections can improve density estimation, though generation tasks—particularly on high-dimensional natural images like CIFAR-10—remain challenging due to the limitations of simple latent models.  

Overall, RPFs offer a low-cost, composable, and theoretically grounded alternative for injective flow modeling, providing a promising building block for future latent-variable and hybrid generative architectures. Future work could explore more expressive latent densities, deeper projections, or hybrid methods to improve generation performance on complex datasets.

\bibliographystyle{unsrt}
\bibliography{neurips}

\newpage
\section*{NeurIPS Paper Checklist}

%%% BEGIN INSTRUCTIONS %%%
The checklist is designed to encourage best practices for responsible machine learning research, addressing issues of reproducibility, transparency, research ethics, and societal impact. Do not remove the checklist: {\bf The papers not including the checklist will be desk rejected.} The checklist should follow the references and follow the (optional) supplemental material.  The checklist does NOT count towards the page
limit. 

Please read the checklist guidelines carefully for information on how to answer these questions. For each question in the checklist:
\begin{itemize}
    \item You should answer \answerYes{}, \answerNo{}, or \answerNA{}.
    \item \answerNA{} means either that the question is Not Applicable for that particular paper or the relevant information is Not Available.
    \item Please provide a short (1–2 sentence) justification right after your answer (even for NA). 
   % \item {\bf The papers not including the checklist will be desk rejected.}
\end{itemize}

{\bf The checklist answers are an integral part of your paper submission.} They are visible to the reviewers, area chairs, senior area chairs, and ethics reviewers. You will be asked to also include it (after eventual revisions) with the final version of your paper, and its final version will be published with the paper.

The reviewers of your paper will be asked to use the checklist as one of the factors in their evaluation. While "\answerYes{}" is generally preferable to "\answerNo{}", it is perfectly acceptable to answer "\answerNo{}" provided a proper justification is given (e.g., "error bars are not reported because it would be too computationally expensive" or "we were unable to find the license for the dataset we used"). In general, answering "\answerNo{}" or "\answerNA{}" is not grounds for rejection. While the questions are phrased in a binary way, we acknowledge that the true answer is often more nuanced, so please just use your best judgment and write a justification to elaborate. All supporting evidence can appear either in the main paper or the supplemental material, provided in appendix. If you answer \answerYes{} to a question, in the justification please point to the section(s) where related material for the question can be found.

\begin{enumerate}

\item {\bf Claims}
    \item[] Question: Do the main claims made in the abstract and introduction accurately reflect the paper's contributions and scope?
    \item[] Answer: \answerYes{} % Replace by \answerYes{}, \answerNo{}, or \answerNA{}.
    \item[] Justification: We derive theory and apply the proposed method with success.

\item {\bf Limitations}
    \item[] Question: Does the paper discuss the limitations of the work performed by the authors?
    \item[] Answer: \answerYes{} % Replace by \answerYes{}, \answerNo{}, or \answerNA{}.
    \item[] Justification: The approach does not greatly improve over other density estimators in every case, which we explain.

\item {\bf Theory assumptions and proofs}
    \item[] Question: For each theoretical result, does the paper provide the full set of assumptions and a complete (and correct) proof?
    \item[] Answer: \answerYes{} % Replace by \answerYes{}, \answerNo{}, or \answerNA{}.
    \item[] Justification: See paper for complete derivation.

    \item {\bf Experimental result reproducibility}
    \item[] Question: Does the paper fully disclose all the information needed to reproduce the main experimental results of the paper to the extent that it affects the main claims and/or conclusions of the paper (regardless of whether the code and data are provided or not)?
    \item[] Answer: \answerYes{} % Replace by \answerYes{}, \answerNo{}, or \answerNA{}.
    \item[] Justification: Yes, see paper for complete derivation.

\item {\bf Open access to data and code}
    \item[] Question: Does the paper provide open access to the data and code, with sufficient instructions to faithfully reproduce the main experimental results, as described in supplemental material?
    \item[] Answer: \answerNo{} % Replace by \answerYes{}, \answerNo{}, or \answerNA{}.
    \item[] Justification: The datasets used are publicly available, and the methods described are straight forward to implement.

\item {\bf Experimental setting/details}
    \item[] Question: Does the paper specify all the training and test details (e.g., data splits, hyperparameters, how they were chosen, type of optimizer, etc.) necessary to understand the results?
    \item[] Answer: \answerYes{} % Replace by \answerYes{}, \answerNo{}, or \answerNA{}.
    \item[] Justification: See Experiments section.

\item {\bf Experiment statistical significance}
    \item[] Question: Does the paper report error bars suitably and correctly defined or other appropriate information about the statistical significance of the experiments?
    \item[] Answer: \answerYes{} % Replace by \answerYes{}, \answerNo{}, or \answerNA{}.
    \item[] Justification: See Experiments section.

\item {\bf Experiments compute resources}
    \item[] Question: For each experiment, does the paper provide sufficient information on the computer resources (type of compute workers, memory, time of execution) needed to reproduce the experiments?
    \item[] Answer: \answerYes{} % Replace by \answerYes{}, \answerNo{}, or \answerNA{}.
    \item[] Justification: See Experiments section.
    
\item {\bf Code of ethics}
    \item[] Question: Does the research conducted in the paper conform, in every respect, with the NeurIPS Code of Ethics \url{https://neurips.cc/public/EthicsGuidelines}?
    \item[] Answer: \answerYes{} % Replace by \answerYes{}, \answerNo{}, or \answerNA{}.
    \item[] Justification: Paper does not violate any of the Code of Ethics.

\item {\bf Broader impacts}
    \item[] Question: Does the paper discuss both potential positive societal impacts and negative societal impacts of the work performed?
    \item[] Answer: \answerNo{} % Replace by \answerYes{}, \answerNo{}, or \answerNA{}.
    \item[] Justification: No particular societal impact besides those standard to generative models.
    
\item {\bf Safeguards}
    \item[] Question: Does the paper describe safeguards that have been put in place for responsible release of data or models that have a high risk for misuse (e.g., pretrained language models, image generators, or scraped datasets)?
    \item[] Answer: \answerNA{} % Replace by \answerYes{}, \answerNo{}, or \answerNA{}.
    \item[] Justification: No models are released, and the datasets used are publicly available.

\item {\bf Licenses for existing assets}
    \item[] Question: Are the creators or original owners of assets (e.g., code, data, models), used in the paper, properly credited and are the license and terms of use explicitly mentioned and properly respected?
    \item[] Answer: \answerYes{} % Replace by \answerYes{}, \answerNo{}, or \answerNA{}.
    \item[] Justification: Benchmarks and datasets are properly cited.

\item {\bf New assets}
    \item[] Question: Are new assets introduced in the paper well documented and is the documentation provided alongside the assets?
    \item[] Answer: \answerNA{} % Replace by \answerYes{}, \answerNo{}, or \answerNA{}.
    \item[] Justification: No new assets released.

\item {\bf Crowdsourcing and research with human subjects}
    \item[] Question: For crowdsourcing experiments and research with human subjects, does the paper include the full text of instructions given to participants and screenshots, if applicable, as well as details about compensation (if any)? 
    \item[] Answer: \answerNA{} % Replace by \answerYes{}, \answerNo{}, or \answerNA{}.
    \item[] Justification: No crowdsourcing was done.

\item {\bf Institutional review board (IRB) approvals or equivalent for research with human subjects}
    \item[] Question: Does the paper describe potential risks incurred by study participants, whether such risks were disclosed to the subjects, and whether Institutional Review Board (IRB) approvals (or an equivalent approval/review based on the requirements of your country or institution) were obtained?
    \item[] Answer: \answerNA{} % Replace by \answerYes{}, \answerNo{}, or \answerNA{}.
    \item[] Justification: No human subjects nor crowdsourcing involved.

\item {\bf Declaration of LLM usage}
    \item[] Question: Does the paper describe the usage of LLMs if it is an important, original, or non-standard component of the core methods in this research? Note that if the LLM is used only for writing, editing, or formatting purposes and does not impact the core methodology, scientific rigorousness, or originality of the research, declaration is not required.
    %this research? 
    \item[] Answer: \answerNA{} % Replace by \answerYes{}, \answerNo{}, or \answerNA{}.
    \item[] Justification: No LLMs used for core research.

\end{enumerate}

\end{document}